# Elderly Fall Detection Using CCTV Cameras under Partial Occlusion of the Subject's Body

SARA KHALILI,[1] HODA MOHAMMADZADE,[2,*] MOHAMMAD MAHDI AHMADI,[1]

[1] Department of Biomedical Engineering, Amirkabir University of Technology, Tehran, Iran

[2] Department of Electrical Engineering, Sharif University of Technology, Tehran, Iran,

*Corresponding author: hoda@sharif.edu

Email Addresses: sarahkhalili89@gmail.com (Sara Khalili), mmahmadi@aut.ac.ir (Mohammad Mahdi Ahmadi)

**Abstract**

One of the possible dangers that older people face in their daily lives is falling. Occlusion is one of the biggest challenges of vision-based fall detection systems and degrades their detection performance considerably. To tackle this problem, we synthesize specifically-designed occluded videos for training fall detection systems using existing datasets. Then, by defining a new cost function, we introduce a framework for weighted training of fall detection models using occluded and un-occluded videos, which can be applied to any learnable fall detection system. Finally, we use both a non-deep and deep model to evaluate the effect of the proposed weighted training method. Experiments show that the proposed method can improve the classification accuracy by 36% for a non-deep model and 55% for a deep model in occlusion conditions. Moreover, it is shown that the proposed training framework can also significantly improve the detection performance of a deep network on normal un-occluded samples.

## 1. INTRODUCTION

The elderly population is increasing dramatically worldwide, and this phenomenon has dire consequences for both the society and the elderly [1]. One of its challenging implications is the increase in the prevalence of frailty in the elderly. Frailty is a state of increased vulnerability, which raises the risk of adverse outcomes, including falls, delirium, and disability [2, 3]. One of the most likely dangers that older people face in their daily lives is falling. Falling is an accident that results in crippling and incapacity, and the danger of this incident reaches a peak when the person is alone and cannot inform others about their condition as this incident may also be accompanied by loss of consciousness [4]. Falling in the elderly endangers their performance and independence; therefore, intelligent and efficient systems in detecting falls are essential.

In recent years, various advances in information and communication technology have been made to help the elderly. One of these technological advances is remote healthcare monitoring, which is evolving rapidly. This technology helps the elderly be independent and gives them the privilege to live in their own homes. Home care is one of the most critical healthcare programs for older people in developed countries, and its primary purpose is to prevent long-term hospitalization and reduce treatment costs [5, 6].

To achieve the primary purpose of human fall detection systems, they need to be trained to distinguish between human falls and other daily life activities (ADL) such as walking, standing, sitting, and lying down. Discrimination between falls and ADL is not easy; because some ADLs have strong similarities with falling, such as lying on the floor or sitting on the floor from a standing position. The training data can be collected from various sensors installed in the environment, such as pressure sensors, floor vibration sensors, infrared sensors, microphones, and cameras. Elderly fall detection systems are divided into three general categories according to their data source: (i) wearable device-based system, (ii) context-aware based system, and (iii) vision-based system [7].

Among these three categories, vision-based systems have recently become more popular due to the ongoing growth of computer vision and deep learning methods. As deep learning and computer vision have been vastly developed in recent years, researchers have used these techniques for automatic fall detection. The cost of implementing vision-based fall detection systems depends on the type and number of cameras in the system. Accordingly, vision-based fall detection methods are classified into three classes: Single-camera systems, multi-camera systems, and depth camera systems. RGB camera based systems are versatile and robust as they can be implemented on previously installed surveillance and security cameras and have better field of view and depth of field as opposed to depth camera systems [8]. On the other hand, single-camera systems have the advantage of simplicity in software and hardware technology; therefore, they are fast, cheap, and suitable for real-time applications [7].

Among the challenges of single-camera vision-based fall detection systems, occlusion is one of the most prominent challenges and degrades their performance to a considerable extent. Occlusion occurs when an object or its main features, which are used to identify it, are not available to the camera [9]. There are two general types of occlusion called partial occlusion and full occlusion. Partial occlusion occurs when some of the main features of the tracked object are hidden from the camera during tracking. Whereas full occlusion occurs when an object being tracked is not visible at all, while it certainly has not left the camera's field of view [9].

In this article, we first evaluate the effect of occlusion on the performance of single-camera vision-based fall detection systems. For this evaluation, we create various meaningful occlusions on the existing fall detection data. We then use specifically-designed occluded videos for training fall detection systems. Finally, by defining a new cost function, we introduce a framework for weighted training of the detection models using occluded and un-occluded videos, which can be applied to any learnable fall detection system.

In classic fall detection methods, a set of various features selected by the system designer is used to distinguish falls from other activities. Despite the promising results reported for most of these methods, their performance is susceptible to deviation from the primary speculations. In contrast, deep learning methods automatically learn features from data. Therefore, they do not work with any initial assumptions and are more suitable for real-life situations given a comprehensive and extensive dataset available for training. On the other hand, implementing deep neural networks on an integrated circuit such as FPGA is expensive due to their massive and complex structure. Thus it will not be cost-effective for the majority of the elderly population.

In conclusion, there is a trade-off between classification accuracy and cost-efficiency for fall detection systems. As a result, both methods are of great importance, and it is necessary to evaluate the effectiveness of the proposed method on both types of systems. Therefore, we select two different models based on classic and deep techniques and evaluate the impact of weighted training on these two models. The first model consists of dynamical Haar cascades as feature extractor and Support Vector Machines (SVM) as classifier. The second model consists of a C3D neural network as feature extractor and SVM as classifier.

This paper is organized as follows. Section II reviews related work. In section III, we investigate the effect of occlusion on the feature space. Section IV gives information on the proposed data augmentation procedure. In Section V, we introduce the weighted training method for occlusion handling. In Section VI, we explain the implementation of this method on two different fall detection algorithms, and in Section VII, we present the obtained results.

## 2. RELATED WORKS

From a general perspective, occlusion handling methods can be classified into three main categories: (I) methods based on the use of multiple cameras, (II) methods based on the use of local features, and (III) methods based on the use of depth information[10]. In the first category, multiple cameras record information from different viewing angles in occlusion conditions. Kung et al. [11] presented a framework in which falls are detected using video surveillance cameras and a three-stream Convolutional Neural Network. In this study, multiple cameras are used alongside a voting system for occlusion handling.

Despite the high accuracy of multiple-camera systems, they are complex and require calibration. Researchers have become more interested in utilizing local features in recent years because of their lower sensitivity to noise and partial occlusion. Numerous works such as [12], [13], [14], and [15] fall under this category. Rougier et al. [12] introduced a new method for detecting falls using the analysis of human shape deformation during a video sequence. In this study, local features such as body edge points are used as key points. Rougier et al. [13] presented another study on fall detection under occlusion conditions, in which a calibrated camera is used to extract the three-dimensional head trajectory in a room. Nguyen et al. [14] proposed a framework for detecting falls using a single camera, in which the center of the mass of the silhouette is used to detect occlusion. In [15], a YOLO neural network is used to estimate the bounding box of the elderly. Then a set of features extracted from the bounding box is used to distinguish falling from other activities. Ge et al. [16] Proposed a human body fall detection algorithm using a fusion hybrid attention mechanism with the aim to increase attention in the process of the fall detection selectivity, and therefore, focusing on not obscured human body parts. Yao et al. [17] introduced a real-time vision-based fall detection method by using geometric features and convolutional neural network. In this study, Gaussian mixture model is utilized to detect the human target and the importance of the head in fall detection is considered by using two different ellipses to represent the head and the torso. Vishnu et al. [18] proposed an approach to efficiently model the spatio-temporal features by training a fall motion mixture model using histogram of optical flow and motion boundary histogram features as well as using the Deep-Sort tracking algorithm for tracking detected multi-objects during occlusions.

Recently, 2D human pose estimation, with or without motion information, has become more popular in occlusion handling. Chen et al. [19] used three relative features, relative coordinates, immediate displacement, and immediate motion orientation, extracted from skeleton sequences and image-based representations of the motion information of the skeleton joints to enhance performance under occlusion. Inturi et al. [20] presented a framework which analyzes the human joint points as prime motion indicators using the AlphaPose pre-trained network. In [21], a tracking human pose estimation method is used to extract the pose features of the human body and solve the occlusion problems in fall detection utilizing motion trajectory. Xie et al. [22] proposed a data fusion method that combine the skeleton keypoints captured from RGB images into fused keypoints to enhance the performance of fall activity classification.

A number of recent studies have used depth information as input to a fall detection system alone or with RGB images such as [23], [24] ,and [25] . Asif et al. [23]

presented a study in which the three-dimensional skeleton is extracted using neural networks. These networks are trained to detect 3D skeletons under partial occlusion. Heitzinger et al. [24] presented a fast 3D detection system which uses an innovative target assignment scheme called broadcasted target assignment (BTA) to improve performance during person-person and person-object occlusions. In [25], a video-based fall detection approach is presented using human poses which first extracts 2D poses from video sequences and then lifts them to 3D poses utilizing a 3D pose estimator.

## 3. Sensitivity of Fall Detection Systems to Occlusion

When a vision-based fall detection systems is trained on un-occluded data, it will perform poorly under occlusion conditions. In fact, occlusion leads to larger intra-class variations and greater inter-class similarities [17]. Fig.1 shows how occlusion affects the distribution of the features of these video sequences. In this figure, the blue and red empty circles represent features of the frames of an un-occluded video sequence. The red circles represent the features of the falling frames, and the blue circles represent the features of the not-falling frames. Moreover, the yellow and green stars indicate the features of the occluded samples that have been synthetically generated from the video sequence, where the yellow stars show the features of the occluded falling frames, and the green stars show the features of the occluded not-falling frames. The detailed procedure of generating these synthetic occlusions is described in Section IV. As shown in Fig.1, occlusion has disordered the feature space and increased intra-class variations.

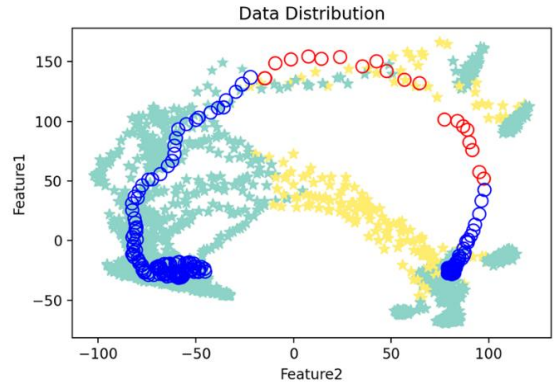

Figure 1. Features of fall (red circles) and not fall (blue circles) in un-occluded video and fall (yellow stars) and not fall (green stars) in occluded videos. These features are extracted from the second fully connected layer in the C3D network. The extracted features originally had 4096 dimensions, which have been reduced to two dimensions by the PCA algorithm. The horizontal axis represents the first feature, and the vertical axis shows the second feature.

## 4. Data Augmentation

It is almost impossible to collect various fall incidents with all of their possible occluded versions. To tackle this dilemma, we propose an augmentation method by generating specific occlusions on videos of the existing datasets (The University of Rzeszow Fall Detection (UR) [26] and Fall Detection (FDD) [27] datasets). In this approach, occlusions are produced based on the coordinates of the joints.

We used the OpenPose human detector network to find the skeleton coordinates. OpenPose is the first real-time multi-person library that jointly detects the human body, hand, facial, and foot keypoints [28]. We utilized the BODY-25 pose output format [28] which can be seen in Fig.2. Moreover, three examples of the detected keypoints in FDD dataset using OpenPose network are shown in Fig.2.

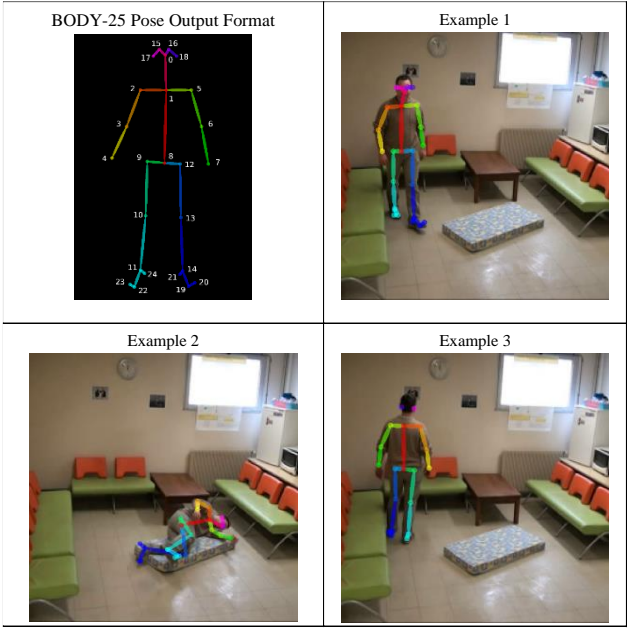

Figure 2. The BODY-25 Pose output format [28] and a few examples of the detected keypoints using OpenPose network

From each original video sequence, ten occluded video sequences are generated using the estimated skeletal coordinates. These occlusions are rectangles with variable length, width, and color. The length and width of each rectangle depend on corresponding joints' coordinates, and its colors are created randomly. These rectangles cover the following body parts, respectively: (1) both legs, (2) head and neck, (3) torso, and both hands, (4) torso, (5) right hand, (6) left hand, (7) right leg, (8) left leg, (9) right side of the body and (10) left side of the body. Note that the rectangle covers the body part throughout the entire video and moves with the subject. Through various experiments, we found out that data augmentation using these moving/dynamic occlusions has a significantly better impact on the detection performance than constant occlusions. Fig.3 shows some examples of the generated occlusions.

## 5. WEIGHTED TRAINING

Using only occluded samples for training may force the model to learn unrealistic features or to ignore useful features in un-occluded parts of the video [29].

Moreover, it is evident that occluded data inevitably has a negative effect on the feature extraction from normal data; therefore, it is necessary to design an effective learning strategy to optimize the model. To this end, we present a weighted method for training fall detection using video sequences. Weighted training using occluded samples has been recently used for face recognition using 2D images [29], but to the best of our knowledge, it has not been used for video classification yet. Moreover, it is important to note that synthesizing effective occlusions in human video sequences is a challenging task because of human movement. It is also important to note that we need an efficient number of synthesized videos with enough occlusions in them to avoid extensive training.

We define the dataset as $I = I_n \cap I_o$ where $I_n$ and $I_o$ represent normal data and its occluded version, respectively. For each sample $x_i \in I$ with label $y_i \in Y$, let's denote its feature $f_i = F(x_i, \theta)$ extracted by function $F$ with parameters $\theta$. The extracted features are then classified by classifier $C$ with parameters $\varphi$ and the predicted label is denoted as $\hat{y} = C(f_i, \varphi)$. The classification cost of the model is calculated as:

$$L = \sum_{i=1}^{N} l(C(f_i, \varphi), y_i) \qquad (1)$$

where $l$ is an arbitrary cost function.

In order to train the model using the two types of samples $x^n \in I_n$ and $x^o \in I_o$, two important points should be considered. The first point is how to resolve the imbalance between the two datasets in training the model. This is because e.g., the number of occluded training samples might be much larger than the number of un-occluded training samples and we do not want the model to be biased toward the occluded samples. The next point is how to control the effect of the occluded samples on the parameters of the model, that is, how to manipulate the feature space to extract the un-occluded features instead of a combination of normal and occluded features [29]. To create the equilibrium, we modify the cost function as:

$$L = L_{x^n \in I_n} + \lambda \frac{n}{o} L_{x^o \in I_o} \quad (2)$$

where $n$ is the number of normal samples, o is the number of occluded samples, and λ determines the weight of the normal and occluded datasets in the model training and has a value between zero and one. In (2), the factor $\frac{n}{o}$ prevents bias toward a large occluded dataset and normalizes the weight factor against the size of original and occluded data. Therefore, the weight is not needed to be tuned when the dataset changes. Moreover, the balance between the normal and occluded features is held using parameter $\lambda$.

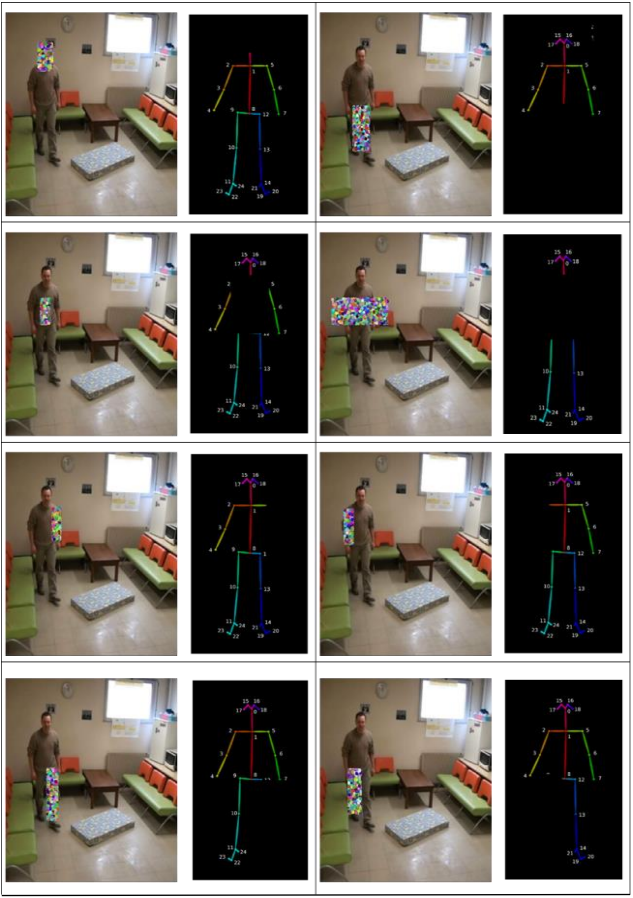

Figure 3. Some examples of the dynamic occlusions

The training framework is illustrated in Fig.4, in which $I_n$ indicates normal video sequences, and $I_o$ indicates occluded video sequences. After extracting the features and performing the classification, the errors are calculated and back propagated to correct the parameters of the model as shown in Fig.4. Classification error for normal data is fully backpropagated, but only a proportion of classification error of occluded data is backpropagated. As a result, the effect of the normal and occluded data on model training can be balanced.

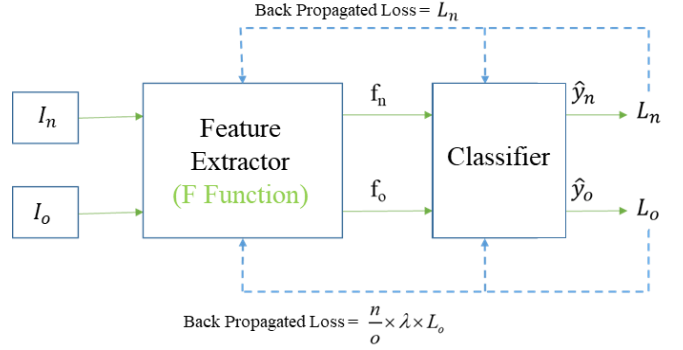

Figure 4. An illustration of the proposed framework. $I_n$ and $I_o$ indicate not-occluded and occluded data and $L_n$ and $L_o$ are the corresponding classification error.

6. WEIGHTED TRAINING OF FALL DETECTION MODELS

To apply the weighted training method on fall detection systems comprehensively, we implement it on two different fall detection methods. The first method is based on dynamic Haar features, and the second is based on deep features.

A. *Haar feature-based fall detection model*

Dynamic Haar features were first proposed for pedestrian detection [30] and have been recently extended for fall detection [31]. In this method falling is distinguished from other activities using motion and appearance filters applied on every four consecutive frames. After feature extraction, 700272 features are obtained using all possible motion and appearance filters for every four consecutive frames. The most distinctive features that can best classify fall incidents from other activities are then selected by the AdaBoost algorithm [31].

In order to find the most distinctive features, we use the weighted training method according to (2). The weight of each sample indicates how important its classification is. We select the top 300 features using the AdaBoost algorithm in the training stage, and with these features, we construct the final feature extractor for the test samples.

After extracting the feature, we use a similar approach for training the classifier. Due to its high accuracy, we use Support Vector Machines method to classify the extracted feature vectors. From a geometric point of view, we can model weighted SVM by moving training samples towards the classifier hyperplane proportionate to their weights ($g_i$) [32]. The formula for the weighted SVM can be defined as:

$$\frac{|f(x_i)|}{\|w\| \times (1+g_i)} = \frac{w^T x + w_0}{\|w\| \times (1+g_i)} \qquad (3)$$

Thus, if the weight of a training sample is zero, its distance to the hyperplane does not change, and as its weight gets larger, its distance becomes smaller [32].

B. *Deep feature based fall detection model*

One of the most successful neural networks in motion detection and activity recognition is C3D network. This network is capable of extracting the temporal and spatial features, useful for body motion detection, human activity localization, and human-scene interaction tasks [33]. In order to apply weighted training according to (2), we define two stages in each iteration of training. In the first stage, a batch of normal video sequences are fed to the network, and their classification error is calculated based on the current network parameters. In the second stage, the occluded samples generated from the same normal samples are fed to the network and their classification error is calculated. The final error is then calculated according to (2), and this error goes back to update the parameters. This process continues until we reach the maximum number of iterations.

7. EXPERIMENTS

The University of Rzeszow Fall Detection (UR) [26] and Fall Detection (FDD) [27] datasets were used to evaluate the proposed method. In addition to these datasets, we recorded 90 non-fall videos and 108 fall videos in various house rooms under occlusion conditions. These videos were recorded using a single camera located in the upper corner of the room. We call this dataset the collected dataset.

In order to perform the experiments, we divided UR and FDD datasets into three sets: training, validation, and test. Sixty percent of the video sequences are randomly selected for the training set, 20% for the validation set and the remaining 20% for the test set. All the frames of each video sequence are entirely placed in either the training, the validation, or the test set. It should be noted that there might be similar environments and people in these sets. We call these training and test sets, the *original* training and test sets, respectively.

In order to evaluate the proposed method, it is vitally essential to generate occlusions that resemble real-life occlusions to a great extent. Therefore, we used objects such as tables, chairs, beds, and other furniture to create occlusions in FDD and UR datasets. It is important to note that these generated occlusions using real-life objects, called realistic occlusion, are only used for validation and test of the detection models and they are different from the dynamic occlusions used for data augmentation. Examples of these realistic occlusions along with the occluding object are shown in Fig.5. We call the test set enriched by their realistically occluded versions, the *occlusion-included* test set. Since the validation set should resemble the samples in the test set, the same procedure that is used for the test set is followed for the validation set too.

The video sequences, as they are, cannot be used as the inputs of deep and non-deep models. We therefore divide each video sequence into consecutive non-overlapping segments. Two parameters are defined here called *sampling-rate* and *stride*. The *sampling-rate* parameter reduces the number of frames. The *stride* parameter specifies the distance between the first frames of two consecutive video segments. Fig.6 depicts the effect of the

*stride* and *sampling-rate* parameters on the number of video segments.

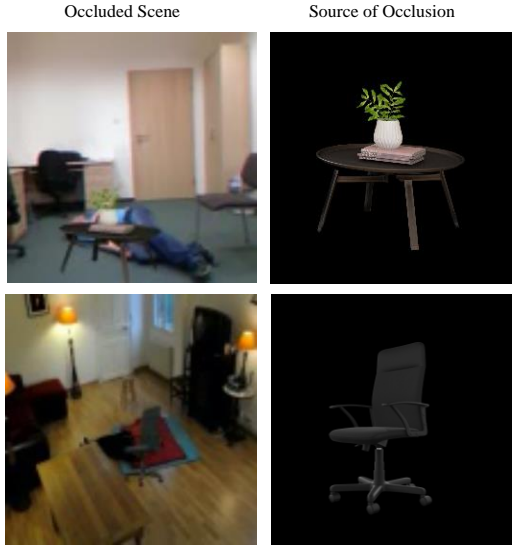

Figure 5. An example of generated realistic occlusions. These occlusions are created using home objects on the UR and FDD datasets.

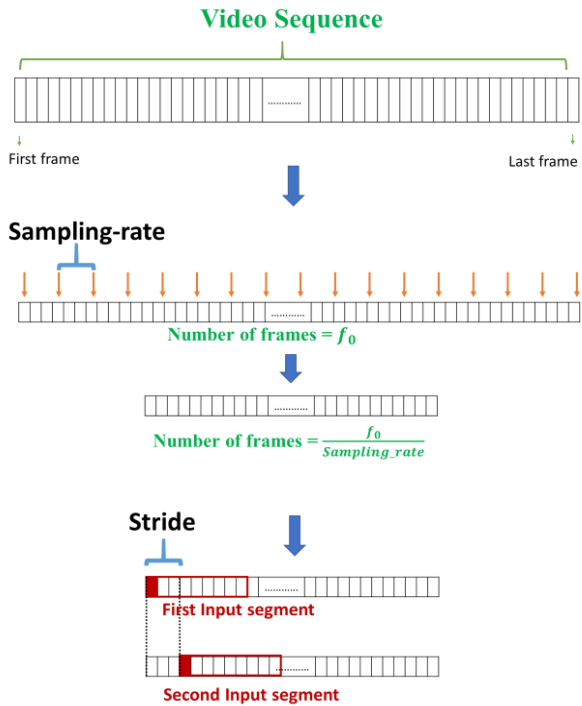

Figure 6. The effect of *stride* and *sampling-rate* parameters on the length of the video sequence. As can be seen, *stride* parameter defines the distance between the two consecutive input patch and *sampling-rate* parameter reduces the number of frames.

For labeling the input segments, we use the following procedure. The initial and final frames within which the falling occurs are annotated in the UR Fall and FDD data sets. The frames before the start of falling and the frames after the end of falling are labeled as non-fall frames, and the frames that come in between are labeled as fall frames. If the input segment is entirely placed in the fall/non-fall section, it is labeled as fall/non-fall. If the input segment is partly positioned in the fall region, it is discarded from the train set. In contrast, such a segment is included in the test set and its label is determined according to the proportion of falling frames to non-falling ones. If the number of falling frames exceeds the non-falling ones, the input segment is labeled as fall and vice versa. Obviously, the segments of the video sequences that do not contain a falling incident are entirely labeled as non-fall.

In order to show the effect of occlusion on fall detection performance, we first experiment with the original training and test sets and then compare the performance with the case that the original training set is used for training but the occlusion-included test set is used as the test set. The results are shown in Table 1 for both the Haar-based and deep based models. As it can be seen, the accuracy drops by 29% for the Haar-based and by 45% for the deep based model when the test set includes occluded videos, showing that the occlusion significantly deteriorates the performance of a fall detection system for both models.

Next, in order to evaluate how much the proposed weighted training model improves the performance when occlusion exists, we perform training using the original training set and the proposed weighted training scheme, and use the occlusion-included test set for testing. The results are shown in Table 1. As can be seen, the proposed method improves the accuracy of Haar-based and deep based models by 27% and 53%, respectively, which shows significant improvement. Similar trends exist for the recall and precision rates. Very interestingly, the performance of the deep-based model has been improved (by 8%) over the original test set (normal data). In other words, the weighted training model has been performed better not only on the occluded data but also on the normal data compared to the non-weighted model. The reason can be that by incorporating the occluded data in the training set, the network is able to extract more robust features, similar to the effect of using the dropout technique in the training of a deep network.

Next, we evaluate the generalization ability of the proposed training method when the training and test sets are from different datasets. For this purpose, we use the fall detection models trained on the original training set and also the one trained using the proposed weighted training scheme, and test on our own collected dataset which include both normal video sequences and their real-life occluded versions. The results are shown in Table 2. As shown in this table, the test samples are from a different dataset, the accuracy of both models drop. The drop is 15% for the Haar-based model and 5% for the deep-based model when original training sets are used. When the weighted training method is used the accuracy of the Haar-based model drops by only 6% and the accuracy of the deep-based model drops by only 3%, showing the generalization ability of the proposed training method.

Table 1. Comparison of the performance of non-deep and deep fall detection models using UR and FDD datasets in three cases: 1) model trained on the original training set and tested on the original test set, 2) model trained on the % original training set and tested on the inclusion-included test set, 3) model trained using the proposed weighted training scheme and test on the inclusion-included test set.

| Haar-based model | | | | |
|---|---|---|---|---|
| Training (number of video sequences) | Test (number of video sequences) | Accuracy | Recall | Precision |
| **Original (2010)** | **Original (600)** | 97% | 91% | 92% |
| **Original (2010)** | **Occlusion-included (1200)** | 68% | 22% | 25% |
| **Weighted training (22110)** | **Occlusion-included (1200)** | 95% | 89% | 84% |
| Deep-based model | | | | |
| **Original (2010)** | **Original (600)** | 91% | 53% | 94% |
| **Original (2010)** | **Occlusion-included (1200)** | 46% | 8% | 72% |
| **Weighted training (22110)** | **Occlusion-included (1200)** | 99% | 96% | 99% |

Table 2. Comparison of the performance of non-deep and deep fall detection models in two cases: 1) model trained on the original training set from UR and FDD datasets and tested on the inclusion-included collected dataset, 2) model trained on the original training set from UR and FDD datasets using the proposed weighted training method and tested on the inclusion-included collected dataset model.

| Haar-based model | | | | |
|---|---|---|---|---|
| Training (number of video sequences) | Test (number of video sequences) | Accuracy | Recall | Precision |
| **Original (2010)** | **Collected dataset (198)** | 53% | 17% | 19% |
| **Weighted training (22110)** | **Collected dataset (198)** | 89% | 83% | 78% |
| Deep-based model | | | | |
| **Original (2010)** | **Collected dataset (198)** | 41% | 19% | 55% |
| **Weighted training (22110)** | **Collected dataset (198)** | 96% | 92% | 89% |

A. Sensitivity Analysis on λ

According to (2), λ creates a balance between the contribution of normal and occluded datasets in the training of a model. Fig.7 shows the effect of λ variation on classification accuracy of the validation data. As can be seen, the best λ value for the Haar-based model is 0.6 and for the deep-based model is 0.3. In addition to that, it is evident that the least accuracy rate is obtained when λ equals zero. In other words, both models perform poorly when occluded samples are eliminated from the training set. However, when occluded samples are fully used in training, a drop occurs in the performance of both models, indicating the importance of balancing the contribution of normal and occluded data in model training.

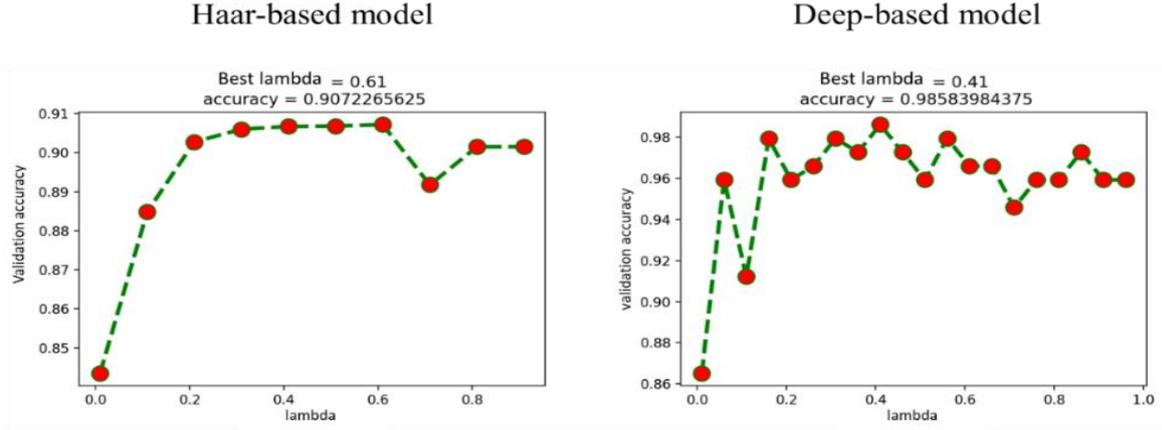

Figure 7. The effect of λ on the performance of two models. The horizontal axis is λ, and the vertical axis shows the classification accuracy of the validation data.

B. *Computational Complexity*

Table 3 presents the training and test time of the Haar-based and deep-based models on UR and FDD datasets. These times are obtained using Google Colaboratory GPU processor with xTesla K80, 2496 CUDA cores, and 12GB GDDR5 VRAM. The test time is reported for one frame of a video. The training time of the Haar-based method is higher than the deep-based method because in the Haar-based method a large number of Haar-like features should be sorted and best of them should be selected. As a result of this efficient selection, the Haar-based method requires less time is the test phase compared to the deep-based method. Specifically, as can be seen in Table 3, the Haar-based model processes a video at 30 frames per second, which therefore can be used for real-time applications.

Table 3. Comparison of the training and test speed of non-deep and deep models using UR and FDD datasets.

|  | Training time | Test time for one sample |
|---|---|---|
| **Haar-based model** | 6.5 hours | 33.371 ms |
| **Deep-based model** | 2.67 hours | 1122.349 ms |

*Ablation Study*

Table 4 depicts a comparison between the augmentation of the training set using the proposed dynamic occlusions versus using the constant occlusions generated in random locations but not moving throughout the video sequence. In this experiment, all other aspects of the data augmentation are identical and the collected dataset is used as the test data. As can be seen, data augmentation using the proposed dynamic occlusions improved the classification accuracy by 12% in Haar-based model and by 23% in deep-based model showing significant improvements for both models.

Table 4. Comparison of the impact of constant and dynamic occlusions on the detection performance of the models.

| Haar-based model | | | | |
|---|---|---|---|---|
| Training (number of video sequences) | Test (number of video sequences) | Accuracy | Recall | Precision |
| **Constant Occlusions (22110)** | **Collected dataset (198)** | 77.4% | 67% | 58% |
| **Dynamic Occlusions (22110)** | **Collected dataset (198)** | 89% | 83% | 78% |
| Deep-based model | | | | |
| **Constant Occlusions (22110)** | **Collected dataset (198)** | 73.8% | 77% | 65% |
| **Dynamic Occlusions (22110)** | **Collected dataset (198)** | 96% | 92% | 89% |

Table 5 reports the effect of the number of the selected features by the AdaBoost algorithm in Haar-based model on the classification accuracy as well as on the computational cost of this model. As can be seen, fewer features lead to lower computational cost, and therefore higher speeds, but lower accuracies. Additionally, the classification accuracy improves by increasing the number of the features but when the number of the features exceeds 300, a 2% drop is visible in the accuracy of the model while the training time significantly jumps, which can be due to the overfitting of the model. Hence the optimum number of the features for the utilized dataset is 300.

Table 5. The effect of the number of selected features on the performance of Haar-based model

| Number of selected features | Accuracy on test data | Duration of training the model | Time for a single output |
|---|---|---|---|
| 10 | 0.72 | 6min | 22ms |
| 100 | 0.88 | 59m36s | 24ms |
| 200 | 0.91 | 1h58m54s | 29ms |
| 300 | 0.95 | 5h37m12s | 33ms |
| 400 | 0.93 | 12h26m36s | 44ms |

## 8. CONCLUSION

This study presented an effective solution to the occlusion problem in single-camera vision-based fall detection systems. In this regard, first, the effect of the occlusion in the feature space was investigated. Then, by the augmentation of the training set through synthetic dynamic occlusions and by using a weighted cost function, we proposed a weighted model training method for fall detection under occlusion. Moreover, we showed that a model trained using the proposed weighted learning method, is not only resistant to occlusion but also becomes better at classifying the normal un-occluded data.